\documentclass[twoside]{article}

\usepackage{aistats2020}
\usepackage{breqn}
\usepackage{amsmath}
\usepackage{amssymb}
\usepackage{amsthm}
\usepackage{algorithm}
\usepackage{algorithmic}
\usepackage{tabulary}
\usepackage{multirow}
\usepackage{graphicx}
\usepackage{float}

\newcolumntype{K}[1]{>{\centering\arraybackslash}p{#1}}
\DeclareMathOperator*{\argmax}{arg\,max}
\DeclareMathOperator*{\argmin}{arg\,min}
\newtheorem{theorem}{Theorem}

\newtheorem{lemma}{Lemma}

\usepackage{xcolor}
\newcommand{\phead}[1]{\textbf{\underline{#1}}}

\newcommand{\tabincell}[2]{\begin{tabular}{@{}#1@{}}#2\end{tabular}}

\begin{document}

%

%

\twocolumn[

\aistatstitle{Triply Robust Off-Policy Evaluation}

\aistatsauthor{ Anqi Liu \And Hao Liu \And  Anima Anandkumar \And Yisong Yue}

\aistatsaddress{Caltech \And  Caltech  \And Caltech  \And Caltech  } ]


\begin{abstract}
    We propose a robust regression approach to off-policy  evaluation (OPE) for contextual bandits.  
    We frame OPE as a covariate-shift problem and  leverage  modern robust regression tools. 
  Ours is a general approach that can  be used to augment any existing OPE method that utilizes the direct method. 
  When augmenting doubly robust methods, we call the resulting method \emph{triply robust}, since we add robustness to the direct method used in doubly robust. 
    We prove upper bounds on the resulting bias and variance, as well as derive novel minimax bounds based on robust minimax analysis for covariate shift.
    Our robust regression method is compatible with deep learning, and is thus applicable to complex OPE settings that require powerful function approximators.  
  Finally, we demonstrate superior empirical performance across the standard OPE benchmarks, especially in the case where the logging policy is unknown and must be estimated from data.
\end{abstract}

\section{Introduction}
Contextual bandits is the online learning setting where a policy repeatedly observes a context, takes an action, and then observes a reward only for the chosen action \cite{langford2007epoch}.  Typical real-world applications include recommender systems \cite{li2010contextual,yue2012hierarchical}, online advertising \cite{tang2013automatic,bottou2013counterfactual}, experiment design \cite{krause2011contextual}, and medical interventions \cite{lei2014actor}.
For settings where online deployments can be costly, an important task is to first  assess a target policy's performance offline, which motivates off-policy evaluation.

Off-policy evaluation (OPE)  is the problem of estimating reward of a target policy from pre-collected historical data generated by some (possibly unknown) logging policy.  The core challenge of OPE is a form of counterfactual reasoning: only the rewards of the actions taken by the logging policy are recorded, and so we must reason about the rewards the target policy would have received despite taking different actions.
To date, there have been many OPE approaches proposed, broadly grouped into three categories: (i) direct methods (DM) that directly regress a target policy's value; (ii) inverse propensity scoring (IPS)  that use importance weights adjustment \cite{horvitz1952generalization,swaminathan2015self}, and (iii) doubly robust (DR) methods that blend  DM and IPS \cite{robins1995semiparametric,bang2005doubly,dudik2011doubly,wang2017optimal,su2019doubly,dudik2014doubly}.

In this paper, we take the perspective of off-policy evaluation as a form of covariate shift \cite{shimodaira2000improving,chen2016robust}. 
Roughly speaking, covariate shift is the problem of modeling a dependent variable, when at test time the generating distribution of the covariates is different than the one used for training.  We will show how to frame OPE as a form of covariate shift, where the dependent variable is the reward model, the covariates are the contexts and actions, and the the generating distributions for the covariates are determined (in part) by the target policy (test time) and the logging policy (training time).
Perhaps surprisingly, thus far there has been little intersection between the covariate shift literature and the OPE literature.

Building upon recent work in deep robust regression under covariate shift \cite{chen2016robust,liu2019robust}, we develop a general framework for augmenting existing OPE methods that utilize a direct method component.  Our contributions are:

\vspace{-0.1in}
\begin{itemize}
\item We show how to frame OPE as a covariate shift problem, and how to leverage modern robust regression tools for tackling covariate shift.
\vspace{-0.05in}
\item We present a general framework for augmenting many OPE methods by using robust regression for the direct method. The resulting DMs can enjoy substantially improved bias and variance compared to their non-augmented counterparts.  When augmenting the DM within a DR method, we call the resulting method \textit{triply robust}, since we add robustness to the direct method used in doubly robust. 
\vspace{-0.05in}
\item We prove bias and variance bounds for our triply robust estimators.  We also  derive novel minimax bounds based on worst-case covariate shift.
\vspace{-0.05in}
\item Our approach is compatible with deep learning, and is thus applicable to complex OPE settings that require powerful function approximators.
\vspace{-0.05in}
\item We demonstrate superior empirical performance across the standard OPE benchmarks, via augmenting several state-of-the-art OPE approaches.  Our approach is particularly beneficial when the logging policy is not known, in which case it can enjoy over 50\% relative error reduction compared to existing state-of-the-art methods.
\end{itemize}


\section{Preliminaries}
\subsection{Off-Policy Evaluation for Contextual Bandits}

In contextual bandit problems, the policy $\pi$ iteratively observes a context $x$, takes an action $a$, and observes a scalar reward $r$.  Assuming the  contexts $x$ are generated iid, the value of a policy can be written as:
\begin{eqnarray}
V^\pi = E_{x\sim P(x), a\sim \pi(x)}\left[r_{x,a}\right],\label{eqn:V}
\end{eqnarray}
where $P(x)$ denotes some exogenous context distribution (e.g., profiles of users), and $a\sim\pi(x)$ denotes the stochasticity of the policy.

In off-policy evaluation (OPE), the goal is to estimate $V^\pi$ offline using pre-collected historical data from some other (possibly unknown) logging policy $p(x)$.  In other words, we assume access to a pre-collected set of $n$ tuples of the form:  $S=\{(x,a,r_{a})\}$, where $x\sim P(x)$, $a\sim p(x)$, and $r_a$ is the observed reward observed for taking action $a$ (we often drop the explicit dependence on $x$ in $r_{x,a}$ when it is clear from context).  We generally do not know $P(x)$ and the reward function, and often also not the logging policy $p(x)$ as well -- they must be estimated using $S$.
Given $S$, the concrete goal of OPE is to compute a reliable estimate $\hat{V}^\pi_S \equiv \hat{V}$ of $V^\pi$ in \eqref{eqn:V} (we typically drop the explicit dependence on $\pi$ and $S$ for brevity).

When designing an effective OPE approach,  the typical considerations are centered around managing the bias-variance tradeoff. Relevant factors include the size of $S$, the degree of overlap between the target policy $\pi$ and the logging policy $p$, and the inherent complexity of estimating the various components $P(x)$, $r(x,a)$, $p(x)$. We next overview several OPE approaches, most of which we can augment using our robust regression framework described in Section \ref{our_method}.  

\subsection{Direct Methods (DM)}
\label{dm}
The simplest class of methods are direct methods (DMs).  DMs aim to directly learn a mapping $\hat{r}$ from  $(x, a)$ to reward $r$, which is essentially a supervised regression problem on $S$ subject to a choice of function class and possibly regularization, e.g., $\hat{r}$ is a neural net trained to minimize:
\begin{eqnarray} \argmin_{\hat{r}} \sum_{(x,a,r_a)\in S}(r_a - \hat{r}(x,a))^2. \ \ \texttt{(DM Training)}\label{dm_train}\end{eqnarray}
Given $\hat{r}$, we can estimate $\hat{V}$ as:
\begin{eqnarray}\hat{V}_{\text{DM}} = \frac{1}{|S|}\sum_{x \in S}E_{a\sim\pi(x)}\left[\hat{r}(x,a)\right].\label{VDM}\end{eqnarray}
DMs are notorious for suffering from a large bias \cite{dudik2011doubly}, because the actions chosen by the target policy $\pi$ are often not chosen by the logging policy $p$, since conventional DM training \eqref{dm_train} is performed over the data collected by the logging policy.
Our key enabling technical insight is to view this issue as a form of covariate shift, and train DMs using robust regression, as described in Section \ref{our_method}.

\subsection{Inverse Propensity Scoring (IPS)}
Inverse Propensity Scoring (IPS) has a rich history in the statistics literature \cite{powell1966weighted,horvitz1952generalization,kang2007demystifying}, and is used in many popular OPE methods.  
Although our framework does not directly augment IPS methods, we provide a brief overview for completeness.

\phead{Vanilla IPS.}
The basic idea is to use important weighting on the entries in $S$ to reflect the relative probabilities of choosing some action $a$ by the target policy $\pi(x)$ versus the logging policy $p(x)$:
\begin{eqnarray}\hat{V}_{IPS} = \frac{1}{|S|} \sum_{(x,a,r_a)\in S}\frac{r_a \pi(a|x)}{\hat{p}(a|x)},\label{VIPS}\end{eqnarray}
where $\pi(a|x)$ is probability of $\pi(x)$ choosing $a$, and $\hat{p}$ is the estimated logging policy (assuming $p$ is not known).  It is known that IPS methods are unbiased but can suffer from high variance if $\pi$ and $p$ diverge strongly in their behavior, due to unstable estimates of the ratio $\pi(a|x)/\hat{p}(a|x)$ \cite{dudik2011doubly}.

\phead{Self-normalized IPS (SnIPS).}
A more recently proposed approach is the
 Self-normalized IPS  estimator \cite{swaminathan2015batch}:
\begin{eqnarray}
\hat{V}_{\text{SnIPS}} =  \frac{\sum_{(x,a,r_a)\in S} r_a\frac{ \pi(a|x)}{\hat{p}(a|x)}} {\sum_{(x,a,r_a)\in S} \frac{ \pi(a|x)}{\hat{p}(a|x)} }.\label{VSnIPS}
\end{eqnarray}
Rather than normalizing by $|S|$ as in vanilla IPS, SnIPS normalizes by the sum of the importance weights.
Even though SnIPS is biased, it tends be more accurate than vanilla IPS when fluctuations in the importance weights dominate  fluctuations in the rewards \cite{swaminathan2015self}. 
It is straightforward for doubly robust methods to use SnIPS as an alternative to vanilla IPS.  

\subsection{Doubly Robust Methods (DR)}
\label{dr_est}
The bulk of recent OPE research for contextual bandits has focused on developing doubly robust estimators, which utilize both DM and IPS as components \cite{dudik2011doubly,dudik2014doubly,wang2017optimal,farajtabar2018more}.  
The basic idea is to balance between the biased but low variance DM and the unbiased but high variance IPS.


\phead{Vanilla DR.} The basic formulation is:
{\small
\begin{eqnarray}
 \hat{V}_{DR} = \hat{V}_{DM} + \frac{1}{|S|}\sum_{(x,a,r_a)\in S}\left[\frac{(r_a - \hat{r}(x,a)){\pi(a|x)}}{\hat{p}(a|x)} \right] , \label{VDR}
\end{eqnarray}
}
\hspace{-5pt}One can also interpret DR estimators as using control variates within an IPS method, albeit traditional control variates tend to be much simpler \cite{magic, veness2011variance}.

Not surprisingly, DR methods depend on  having a good DM or a good IPS. For instance, when one of IPS or DM is unbiased, DR is guaranteed to be unbiased~\cite{dudik2011doubly}. It has also been shown that the variance of DR mainly comes from the IPS term \cite{dudik2011doubly}. Moreover,  when IPS is not accurate or has high variance, an inaccurate DM can have its error compounded within a DR beyond just using the DM alone. As such, a large body of follow up work has focused on how  develop advanced DR methods that mitigate the damaging effects of variance or extreme probabilities from the IPS component. 

\phead{SWITCH.} The SWITCH estimator 
extends vanilla DR by introducing weight clipping \cite{wang2017optimal}.  SWITCH uses vanilla DR unless the importance weight is too large, in which case it only uses the DM.\footnote{There is a version of SWITCH that switches from IPS to DM, rather than from DR to DM.  We omit that version since it typically performs worse.}  
The intuition is to avoid using the IPS term (and thus reduce to only using DM) if the extreme importance weights are harming the effectiveness of DR.
The estimator can be written as:

\vspace{-17pt}
\begin{small}
\begin{align}
&\hat{V}_{\text{SWITCH}} =\notag\\
&  \frac{1}{|S|}\sum_{(x,a,r_a)\in S} \left[\left(\frac{(r_a - \hat{r}(x, a)){\pi(a|x)}}{\hat{p}(a|x)} + \hat{r}_{\pi}(x) \right)\mathbf{1}(w_a \leq \tau) \right] \notag\\ 
& + \frac{1}{|S|}\sum_{x \in S}E_{a\sim\pi(x)}\left[\hat{r}(x,a) \mathbf{1} (w_a>\tau)\right],\label{VSWITCH}
\end{align}
\end{small}
\hspace{-4pt}where $\hat{r}_\pi(x) = E_{a\sim\pi(x)}\hat{r}(x,a)$, $\tau$ is the threshold parameter for switching, and
$w_a = \frac{ \pi(a|x)}{\hat{p}(a|x)}$.  This estimator's performance highly depends on the tuning of the parameter of the weight clipping threshold. 

\phead{DR-Shrinkage.} 
DR with Shrinkage extends vanilla DR by shrinking the IPS term to obtain a better bias-variance trade-off in finite samples \cite{su2019doubly}:
\begin{eqnarray}
\hat{V}_{DRs} = \hat{V}_{DM} + \frac{1}{|S|} \sum_{(x,a,r_a)\in S} \hat{w}(x,a) (r_a-\hat{r}(x, a)),\label{VDRS}
\end{eqnarray}
where ${\hat{w}}:\mathcal{X}
\times \mathcal{A} \to R^{+}$ is a weight mapping found by hard threshold or optimizing a sharp bound on
MSE. In a situation analogous to SWITCH, the performance of DR-Shrinkage is highly dependent on the being able to find a good weight mapping.


\textbf{Towards Triply Robust.}
Perhaps surprisingly, not much work has been done on directly minimizing the bias of DMs.  Instead, recent research has largely focused on designing DR methods that more carefully balance between the IPS and DM components, in order to control for the variance of IPS.\footnote{This methodological focus is also present in research on OPE methods for the RL setting, e.g.,  \cite{jiang2015doubly,magic,farajtabar2018more}. Extending our framework to the RL setting is a natural direction for future work.}
In Section \ref{our_method}, we propose a complementary line of research in leveraging robust regression methods to train DMs, which can then be seamlessly integrated in most DR approaches to arrive at their triply robust counterparts.


\section{Robust Regression for OPE}
\label{our_method}

\subsection{Off-Policy Evaluation as Covariate Shift} 

Covariate shift refers to the distribution shift caused only by the input distribution, while the conditional output distribution remains unchanged \cite{shimodaira2000improving,chen2016robust}. Assuming the logging data is sampled from a joint distribution $P(x, a, r)$, our goal is to accurately estimate the conditional reward distribution $P(r|x, a)$. The $\hat{r}(x,a)$ estimator described in Section \ref{dm} would then be re-defined as the expected value of this reward distribution, rather than using vanilla supervised learning as in \eqref{dm_train}.  Given such a $\hat{r}$, it is straightforward to incorporate it into a direct method such as \eqref{VDM}.

Covariate shift arises because the covariates to the reward model $P(r|x,a)$, in particular the action $a$, experience distribution shift between training and testing.  
The joint distribution over the covariates can be written as $P(x,a) = P(x)P(a|x)$.  The generating distribution for contexts, $P(x)$ is exogenous and fixed (and thus does not contribute to covariate shift).  On the other hand, the conditional action distribution $P(a|x)$ varies depending whether it corresponds to the target evaluation policy or the logging policy.  We explicitly deal with this shift when estimating a reward model from logging data using robust regression. 



Existing methods for dealing with covariate shift typically employ density ratio estimators, which can be very challenging in high-dimensional settings.  In our setting, the contexts $x$ can be very high dimensional, but the actions $a$ are typically low dimensional.  However, since $P(x)$ does not experience distribution drift, then we only need to employ density ratio estimators for $P(a|x)$, which is much easier to do.  As a consequence, OPE, once properly framed, actually reduces to a relatively simple covariate shift problem.

\subsection{Deep Robust Regression}

We now present a deep robust regression approach for off-policy evaluation. The naming of ``robust'' originates from a line of research in statistics on robust estimation under distribution drift \cite{shimodaira2000improving}.  The high level goal is to estimate a reward model $\hat{P}(r|x,a)$ that is robust to the ``most surprising'' distribution shift that can occur, which can be formulated using a minimax objective.   We build upon modern tools for deep robust regression under covariate shift  \cite{liu2019robust}.

\textbf{Relative Loss.}
For technical reasons, it is convenient to design a relative loss function defined as the difference in conditional log-loss between an estimator $\hat{P}(r| x,a)$ and a baseline conditional distribution $P_0(r| x,a)$ on the target data distribution $P(x)\pi(a|x)P(r|x,a)$.  This loss essentially measures the amount of expected ``surprise" in modeling true data distribution $P(x)\pi(a|x)P(r|x, a)$ that comes from $P(x)\pi(a|x)\hat{P}(r| x, a)$ instead of $P(x)\pi(a|x)P_0(r|x, a)$:
 \begin{eqnarray}
     \mathcal{L}:=E_{a\sim\pi(a|x),r\sim P(r|x, a)}\left[
    -\log\frac{\hat{P}(R| X, A)}{P_0(R|  X, A)}
    \right]. \label{eq:relativeloss}
 \end{eqnarray}
 The choice of $P_0$ is straightforward in most applications, and we typically use a Gaussian distribution. 
 
\textbf{Quantifying Allowable $P(r|x,a)$.} The next step is to quantify the allowable conditional distribution that we aim to be robust against.  
We do so by creating a constrained set $\Gamma$ of allowable $P(r|x, a)$ that are consistent with data statistics from covariate distributions $P(x,a)=P(x)P(a|x)$:
\begin{align}
&\Gamma:=\{P(r|x, a) |\notag\\
&\quad \quad \quad E_{a\sim p(a|x), r\sim P(r|x, a)}[f(x, a, r)] - {\bf c}| \le \eta \},\label{eq:constraintWithc}
\end{align}
where ${\bf c} = \frac{1}{n}\sum^{n}_{i = n}f(x_i, a_i, r_i)$ is a vector of statistics measured from the logging data,  and $\eta$ is a hyperparameter. Note that $\Gamma$ is defined on the logging data distribution, while $\mathcal{L}$ is defined on evaluation data distribution.  The crux of this definition lies in the specific instantiation of $f$, which we discuss next.

\textbf{Interpreting $f$}.
Originally developed for linear function classes \cite{chen2016robust},  $f$ is typically instantiated as linear or higher-moment statistics, which in \eqref{eq:constraintWithc} correspond exactly to quantifying the allowable distribution drift via the drift in the sufficient statistics.
This interpretation is somewhat less clear when extending to deep neural networks (although the bias/variance and minimax bounds described in Sections \ref{analysis1} \& \ref{analysis2} are still valid).  
In the deep neural net case, we define $f$ as the top hidden layer, which can be estimated  end-to-end during training \cite{liu2019robust}. In other words, we directly learn the sufficient statistics to characterize distribution shift.

\textbf{Minimax Objective.}
Our learning goal is to find a regression model that is robust to the ``most surprising" conditional distribution $P(r|x,a )$ that can arise from logging data distribution but still consistent with evaluation data distribution under covariate shifts: 
\begin{eqnarray}
    \min_{\hat{P}(r|x, a)}\max_{P(r|x, a)\in \Gamma} \mathcal{L}.\label{eq:minimax}
    \end{eqnarray}
By using relative loss \eqref{eq:relativeloss} with $P_0=\mathcal{N}(\mu_0,\sigma_0^2)$, along with the constraint formulation in \eqref{eq:constraintWithc}, the solution to \eqref{eq:minimax} takes the form of a conditional Gaussian distribution $P_{\rho} \sim N(\mu(x, a, \rho), \sigma^2(x, a, \rho))$:

\vspace{-14pt}
{\small
\begin{align}
    \sigma^2(x, a, \rho)& = \left(2 \frac{p(a|x)}{\pi(a|x)} \rho_r + \sigma_0^{-2}\right)^{-1}, \label{DMR}\\
    \mu(x,a,\rho) &= \sigma^2(x,a,\rho)\left(-2\frac{p(a|x)}{\pi(a|x)} \rho_{xr} 
    f(x, a, \theta)  + \mu_0\sigma_0^{-2}\right),\notag 
\end{align}
}
\hspace{-6pt}where $\rho$ is a matrix: $\rho = \begin{bmatrix} \rho_r &\rho_{xr}\\ \rho_{xr} &\rho_{xa}  \end{bmatrix}$,  $\mathcal{N}(\mu_0, \sigma^2_0)$ is the base distribution $P_0$, and $f(x, a, \theta)$ is the top hidden layer of a neural net.   
A detailed derivation is available in Appendix \ref{app_a}. 

{\bf Learning Algorithm.} Another technical convenience of this formulation is that, during learning, we do not explicitly consider  $\Gamma$, since it is included in the KKT conditions at optimality (see the appendix).  We can thus employ standard gradient-based learning, as summarized in Algorithm \ref{alg:sgd}.

{\bf The role of density ratio $\frac{p(a|x)}{\pi(a|x)}$:} The density ratio $\frac{p(a|x)}{\pi(a|x)}$ corresponds to the logging policy over evaluation policy. The intuition can be interpreted assuming both logging and evaluation policy are stochastic policies. For a certain action, if the probability under the logging policy is different from the one under logging policy, we should adjust our prediction uncertainty. Especially, when an action is very probable under the evaluation policy but improbable under the logging policy, the estimator tends to be less certain and depends more on the base distribution $P_0(R|  X, A)$.  

{\bf The role of base distribution $P_0(R|  X, A)$:} The base distribution provides prior knowledge about the ``default" conditional reward distribution choice when the logging policy and the evaluation policy is totally distinct on certain actions, which is when $\frac{p(a|x)}{\pi(a|x)}$ is close to 0. For OPE, it is reasonable to choose a Gaussian distribution with mean equals to the $\frac{r_{\text{max}}+ r_{\text{min}}}{2}$, where $r_{\text{max}}$ and $r_{\text{min}}$ define the range the reward values can take, assuming we do not have more informative knowledge about the reward.


     \begin{algorithm}[t]
        \caption{Stochastic Gradient Descent for Deep Robust Regression under Covariate Shift}
         \label{alg:sgd}
         \begin{small}
        \begin{algorithmic}
          \STATE{\bfseries Input}: Training data points $\{( x_i, a_i, r_i)\}$, logging policy $p(a_i|x_i)$, evaluation policy = $\pi(a_i|x_i)$,  DNN $f(x, a)$ with initialization,  DNN SGD optimizer $Opt$, learning rate $\gamma$, regularization $\eta$, epoch number $T$.
            \STATE $\Theta \leftarrow$ random initialization, epoch $= 0$
            \STATE {\bfseries While} epoch $< T$
            \STATE \qquad {\bfseries For} each mini-batch
            \STATE \qquad\qquad Obtain top hidden layer $f(x, a, \theta)$ 
            \STATE \qquad\qquad Compute $\mu(x, a, \rho)$ and $\sigma^2(x, a, \rho)$ (Eq. \ref{DMR})
            \STATE \qquad\qquad Compute gradients for $\rho$ (details in appendix Eq. \ref{eq:gradients1} and Eq. \ref{eq:gradients2}.)
            \STATE \qquad\qquad Gradient descent on $\rho_r$ and $\rho_{xr}$ with regularization $\eta$  
            \STATE \qquad\qquad Back-Propagate through networks.
            \STATE \qquad\qquad SGD using $Opt(f, \gamma).step()$
            \STATE {\bfseries Output}: Trained NN and $\rho$
        \end{algorithmic}
        \end{small}
      \end{algorithm}

\subsection{Triply Robust Off-Policy Evaluation}
\label{tr}

We now overview how our robust regression approach can be used to augment many existing OPE methods that utilize a direct method component. 

\phead{DM-R}. We can augment vanilla DMs by plugging in mean estimates from robust regression to obtain:
\begin{eqnarray}
    \hat{V}_{\text{DM-R}} = \frac{1}{|S|}\sum_{x \in S}E_{a\sim\pi(x)}\left[\mu(x,a, \rho)\right].\label{VDMR}
\end{eqnarray}
\phead{TR}. The triply robust method augments DR by augmenting the DM component. For simplicity, we use $\hat{\mu}_a$ and $\hat{\mu}_{\pi}$ to represent the mean prediction from robust regression on logging policy and evaluation policy.

\vspace{-10pt}
\begin{small}
\begin{dmath}
    \hat{V}_{TR} = \frac{1}{|S|}\sum_{(x,a,r_a)\in S}\left[\frac{(r_a - \hat{\mu}_a)\pi(a|x)}{\hat{p}(a|x)} \right] + \hat{V}_{DM-R}. \label{VTR}
\end{dmath}
\end{small}
Similar with DR, TR benefit from controlling the variance of IPS by using SnIPS or using SWITCH and Shrinkage based on (optimized) thresholds. We list these methods below. 

\phead{SnTR}. Using Self-normalized IPS in the first term of TR, we obtain: 
    \begin{align}
        &\hat{V}_{\text{SnTR}} =  \frac{\frac{1}{|S|}\sum_{(x,a,r_a)\in S}\frac{(r_a - \hat{\mu}_a)\pi(a|x)}{\hat{p}(a|x) } }{\frac{1}{|S|}\sum_{(x,a,r_a)\in S}\frac{\pi(a|x)}{\hat{p}(a|x) }}
        + \hat{V}_{\text{DM-R}} .\label{VSnTR}
    \end{align}
    
\phead{TR-SWITCH}. As in SWITCH, we switch from TR to DM-R at a certain threshold $\tau$, and
$w_a = \frac{ \pi(a|x)}{\hat{p}(a|x)}$:

\vspace{-15pt}
    {\small 
    \begin{align}
        &\hat{V}_{\text{TR-SWITCH}} \notag\\
        &\ =  \frac{1}{|S|}\sum_{(x,a,r_a)\in S} \left[\left(\frac{(r_a - \hat{\mu}_a){\pi(a|x)}}{\hat{p}(a|x)} + \hat{\mu}_{\pi} \right)\mathbf{1}(w_a \leq \tau) \right] \notag\\ 
        &\ \quad \ + \frac{1}{|S|} \sum_{(x,a,r_a)\in S}\hat{\mu}_{\pi} \mathbf{1} (w_a>\tau).\label{VTRSWITCH}
    \end{align}
    }
\hspace{-7pt}\phead{TR-Shrinkage}. As in DR-Shrinkage, we use a customized importance weight $\hat{w}(x,a)$ for the first term of TR, which needs to be tuned or optimized carefully:
   \begin{eqnarray}
    \hat{V}_{\text{TRs}} = \hat{V}_{\text{DM-R}} + \frac{1}{|S|} \sum_{(x,a,r_a)\in S} \hat{w}(x,a) (r_a-\hat{\mu}_a).\label{VTRS}
    \end{eqnarray}

\subsection{Bias and Variance Analysis}
\label{analysis1}
Our analysis connects learning generalization bound of direct method and bias and variance analysis in doubly robust to obtain upper bounds for both bias and variance analysis in the Triply Robust. We first denote $\epsilon$ as the generalization error $E_{a\sim \pi(a|x), r\sim P(r|x, a)}[(r - \hat{\mu}_a)]$ upper bound that is given in Theorem 1 in~\cite{liu2019robust}. We refer to appendix \ref{app_b} for a detailed restatement of the bound in the off-policy evaluation setting. 
\begin{theorem}
\label{thm:bias}
The bias of triply robust is bounded by the following with probability at least $1-\delta$:
\begin{align}
 &| E_{a\sim \pi(a|x), r\sim P(r|x,a)}[\hat{V}_{TR}] - V^{\pi}] |\notag\\
  &\le W\eta_1/l  + \epsilon + O(\sqrt{W\log (1/\delta)/n}),
\end{align}
where $W$ is the upper bound of $\frac{\pi(a|x)}{p(a|x)}$, and $n$ is the number of data samples.
\end{theorem}

\begin{theorem}
\label{thm:variance}
The variance of triply robust method is bounded by the following with probability at least $1-\delta$:
\begin{align}
    &VAR_{a\sim \pi(a|x), r\sim P(r|x,a)}[\hat{V}_{TR}] \notag\\
    & \quad \quad \le 2W^2\eta_2 + 2W^2(2WB + \frac{1}{\sigma_0^2})^{-1}\notag\\
    &  \quad  \quad \quad+ O(W^2\sqrt{\log (1/\delta)/n}) + 2\epsilon^2
\end{align}
where $W$ is the upper bound of $\frac{\pi(a|x)}{p(a|x)}$, $B$ is the upper bound of model parameter $\rho_r$, and $n$ is the number of data samples.
\end{theorem}
To interpret this two bounds, both the bias and variance is upper bounded by a combination (1) moments of generalization error of robust regression on evaluation data $\epsilon$ and (2)the constraint violations in the logging data that is related with weighted $\eta_1$ and $\eta_2$ by the IPS. Therefore, this shows a good direct method could help reduce the bias and variance of TR.

\subsection{Minimax Analysis}
\label{analysis2}
Minimax analysis provides insights about the best possible performance among all the statistical procedures under the worst case behavior of a method. It has been shown in a general case under the multi-armed bandit case ~\cite{li2015toward} and contextual bandit setting~\cite{wang2017optimal}. In our case, instead of focusing on general max mean and variance constraints, we utilize the data dependant constraints as in \eqref{eq:constraints} (in Appendix \ref{app_c}) and obtain a data dependent minimax analysis on DM-R.

Recall that under the robust regression framework, slack terms like $\eta_1$ and $\eta_2$ correspond with the regularization in parameter optimization~\cite{chen2016robust}. So we assume they are bounded. Recall $f$ is the representation of covariates (x, a) in robust regression.
\begin{theorem}
\label{thm:minimax}
Assuming $w(x, a) = \frac{\pi(a|x)}{p(a|x)} \le \infty$, $w$ is independent with $r$, define $\mathcal{S}(\eta_1, \eta_2, f)$ as the set of distributions that satisfy \eqref{eq:constraints} (in Appendix \ref{app_c}), the minimax risk $\mathcal{R}(P, p, \pi, \eta_1, \eta_2, f)$ of off-policy evaluation over the class  $\mathcal{S}(\eta_1, \eta_2, f)$, which is defined as $\inf_{\hat{r}} \sup_{D(r|x, a) \in \mathcal{S}(\eta_1, \eta_2, f)} E_{a\sim \pi(a|x), r\sim D(r|x,a)}[(\hat{r} - r(x, a))^2]$ satisfies the lower bound:
\begin{align}
    &\min \Big\{ \frac{w^2\eta^2_2}{64el^2},\notag\\
    &\quad\quad\quad \frac{(-4E_{p}[wr] + \sqrt{16 E_{p}[wr] ^2+ 8w^2(n+2)\eta_1})^2}{128e(n+2)^2}
    \Big\},\notag
\end{align}
where we abuse the notation a little and use $E_p$ to represent $E_{a\sim p(a|x),r\sim P(r|x,a)}$ in the expectation in the second term, $l$ is the lower bound of $f(x, a)$, n is the number of data samples.
\end{theorem}
The minimax lower bound of DM-R is the minimum of two terms that are related with $\eta_1$ and $\eta_2$ respectively. Unlike other minimax risk analysis, our bound is not related with the upper bound of variance $\sigma_{\text{max}}^2$ but is closely related with expectation of weighted reward in the logging data distribution. This is due to the fact that constraints in \eqref{eq:constraints}, defines the relation between mean and variance of the resulting conditional reward distribution, given fixed $\eta_1$ and $\eta_2$. 




\section{Related Work}



\textbf{Advances in Off-Policy Evaluation and Learning:}
Modern off-policy evaluation methods use powerful tools like deep learning to deal with data in large dimensionality and volume, and can also be used within off-policy learning approaches. BanditNet \cite{joachims2018deep} provides a counterfactual risk minimization
approach for training deep networks using an equivalent empirical risk estimator with variance regularization. We use deep robust regression for off-policy evaluation and is also compatible with a number of off-policy optimization methods.

Off-policy evaluation has been studied in scenarios other than traditional contextual bandits setup, such as slate recommendation \cite{swaminathan2017off}, where key challenge is that the number of possible lists (called slates) is combinatorially large. Off-policy evaluation has been a key challenge in reinforcement learning \cite{magic,fqe,xie2019optimal,jiang2015doubly}. It has also been considered in the setup where there are multiple logging policies \cite{he2019off}. 

{\bf Causal Inference:} Off-policy evaluation is connected closely with causal inference\cite{athey2015machine}. A key problem for evaluating the individual treatment effect (ITE) and average treatment effect (ATE) is the evaluation of a counterfactual policy. Methods from domain adaptation and deep representation learning 
\cite{johansson2016learning} has been applied in this area, but still falls in the sample re-weighting category. There has also been work on using causal models to achieve better off-policy evaluation result \cite{oberst2019counterfactual}.

{\bf Robust Regression and Covariate Shift:} 
Importance weighting methods  is the common choice for regression under the distribution shift\cite{shimodaira2000improving, sugiyama2007covariate}. However, though being asymptotically unbiased, it suffers from the high variance. Recently developed robust covariate shift methods take a worst-case approach, constructing a predictor
that (approximately) matches training data statistics, but is otherwise the most uncertain on the testing
distribution. These methods were
built by minimizing the worst-case expected target log loss and obtain a parametric form of the predicted output labels’ probability distributions \cite{chen2016robust,liu2017robust}. We are the first to use these types of robust regression methods for off-policy evaluation and are able to construct better direct method and further improve doubly robust estimator.

\section{Experiments}

\subsection{Setup}

We validate our framework across the standard OPE benchmarks considered in prior work.
In particular, we use several UCI datasets as well as CIFAR10, where we convert the multi-class classification problem to contextual bandits with binary reward, following \cite{dudik2011doubly}. Table \ref{tab:data} includes detailed information for datasets we used in the experiments.For each experiment, we first separate the data into training and testing in a 60\% to 40\% ratio. We then use the fully observed training data to train a classifier that would serve as the evaluation policy in the testing. We use certain logging policy to sample an action for each context, which is one of the class labels in our case, to serve as our training data. We use the same logging policy to generate the testing data. In testing, we first evaluate the ground truth of evaluation policy, which is the classification error of the pretrained classifier, and then compare with the off-policy evaluation methods in RMSE and standard deviation. More experimental details are in Appendix \ref{app_e}.
\vspace{-5mm}
\begin{table}[htb]
    \caption{Dataset description for bandit simulation.}
    \centering
    \begin{tabular}{|c|c|c|c|}
         \hline
         Datasets & \#Dimensions & \#Samples & \#Classes\\
         \hline
         vehicle &  18&  946& 4 \\
         \hline
         optdigits& 64 & 5620 &10\\
         \hline
         letter  & 16 & 20000 & 26\\
         \hline  
         CIFAR10&  3072  & 60000 & 10 \\
         \hline
    \end{tabular}
    \label{tab:data}
\end{table}

\vspace{-5mm}

\textbf{Logging Policies.} A nice property of multiclass classification to contextual bandits conversion~\cite{wang2017optimal,dudik2011doubly} is we can control the logging policy to sample training data. Therefore,  to cover various logging policies, our logging policy is trained using a subsampled dataset that is potentially biased. The greater the bias, the more probable there exist extreme densities in the logging policy and the variance of IPS is larger.

We also investigate 
the case of an unknown logging policy, which is  more challenging.  We use a  classification method that optimizes the logloss objective and produces probabilities for each class as the logging policy estimation.    
 


\begin{table}[t]
    \centering
        \caption{Main experimental comparison results, using \textbf{(top)} known and uniform logging policy; \textbf{(middle)} known and high-variance logging policy; and \textbf{(bottom)} unknown logging policy estimated from data. Showing best performing methods in DM/IPS/DR family and DM-R/TR family with their performance in RMSE mean and standard deviation (in parentheses) over 20 repeated experiments. Here we use $DR_s$ and $TR_s$ to represent DR-Shrinkage and TR-Shrinkage.
        We see that the best TR method generally outperforms the best baseline method, especially when the logging policy is unknown.
        }
        \begin{small}
        \begin{tabular}{|K{1.2cm}|K{2.9cm}|K{2.9cm}|}
         \hline
         \textsc{Data} & \tabincell{c}{Best in \\ DM/IPS/DR } & \tabincell{c}{Best in \\ DM-R/TR }\\
         \hline
         {vehicle}&  $DR_s$: 0.028(0.024)& $TR_s$: 0.026 (0.023) \\
         \hline
         {optdigits} & DR: 0.046 (0.030)& TR: 0.045 (0.028) \\
         \hline
         {letter} &DR: 0.021 (0.021) &TR: 0.019(0.019) \\
         \hline
         {CIFAR10} & DR: 0.012 (0.0092) &TR: 0.011(0.0088) \\
         \hline
             \end{tabular}
             
             \medskip
             
        \begin{tabular}{|K{1.2cm}|K{2.9cm}|K{2.9cm}|}
         \hline
\textsc{Data} & \tabincell{c}{Best in \\ DM/IPS/DR } & \tabincell{c}{Best in \\ DM-R/TR }\\        \hline
         {vehicle}& DM: 0.070($<$10e-6)& DM-R: 0.0076($<$10e-6) \\
         \hline
         {optdigits} & DR: 0.21 (0.20)& TR: 0.13 (0.13) \\
         \hline
         {letter}& DR: 0.061 (0.061) &TR: 0.040(0.028) \\
         \hline
         {CIFAR10}&DR: 0.015 (0.0060) &TR: 0.012(0.0050) \\
         \hline
    \end{tabular}
    
    \medskip
    
        \hspace{2.5pt}
       \begin{tabular}{|K{1.2cm}|K{2.9cm}|K{2.9cm}|}         
       \hline
\textsc{Data} & \tabincell{c}{Best in \\ DM/IPS/DR } & \tabincell{c}{Best in \\ DM-R/TR }\\        \hline
         {vehicle}& IPS: 0.21 (0.089)& $TR_s$: 0.18 (0.013)\\
         \hline
         {optdigits} & DR: 0.53 (0.025)& TR: 0.47 (0.022)\\
         \hline
         {letter}  &DR: 0.033 (0.016)& TR: 0.022 (0.016)\\
         \hline
         {CIFAR10}  &DR: 0.070 (0.012) &TR: 0.033(0.012) \\
         \hline
    \end{tabular}
    \end{small}
    \label{tab:main_experiments}
\end{table}

\begin{figure*}[htbp]
\begin{tabular}{cc}
      
    \includegraphics[width=0.48\textwidth]{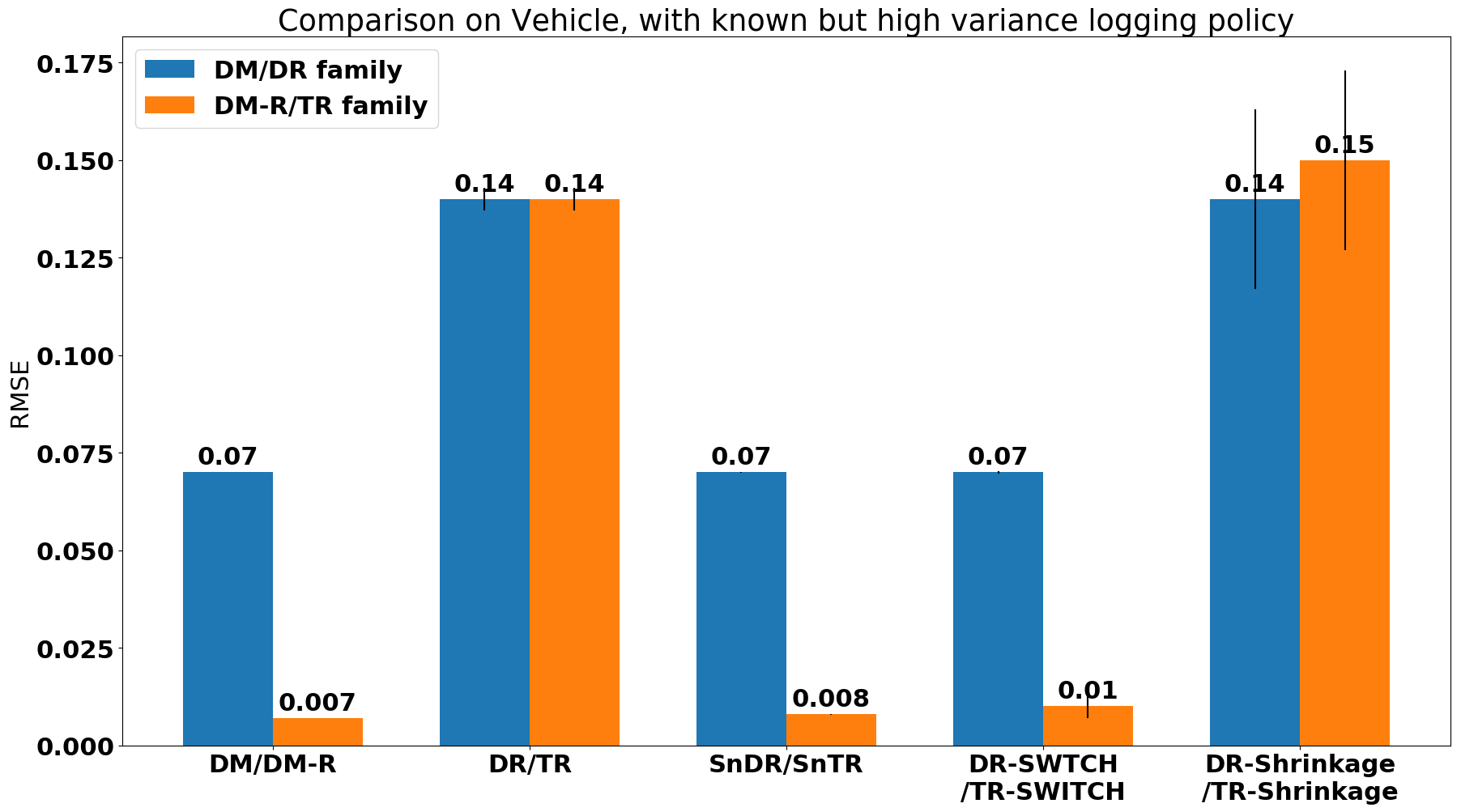}&
     \includegraphics[width=0.48\textwidth]{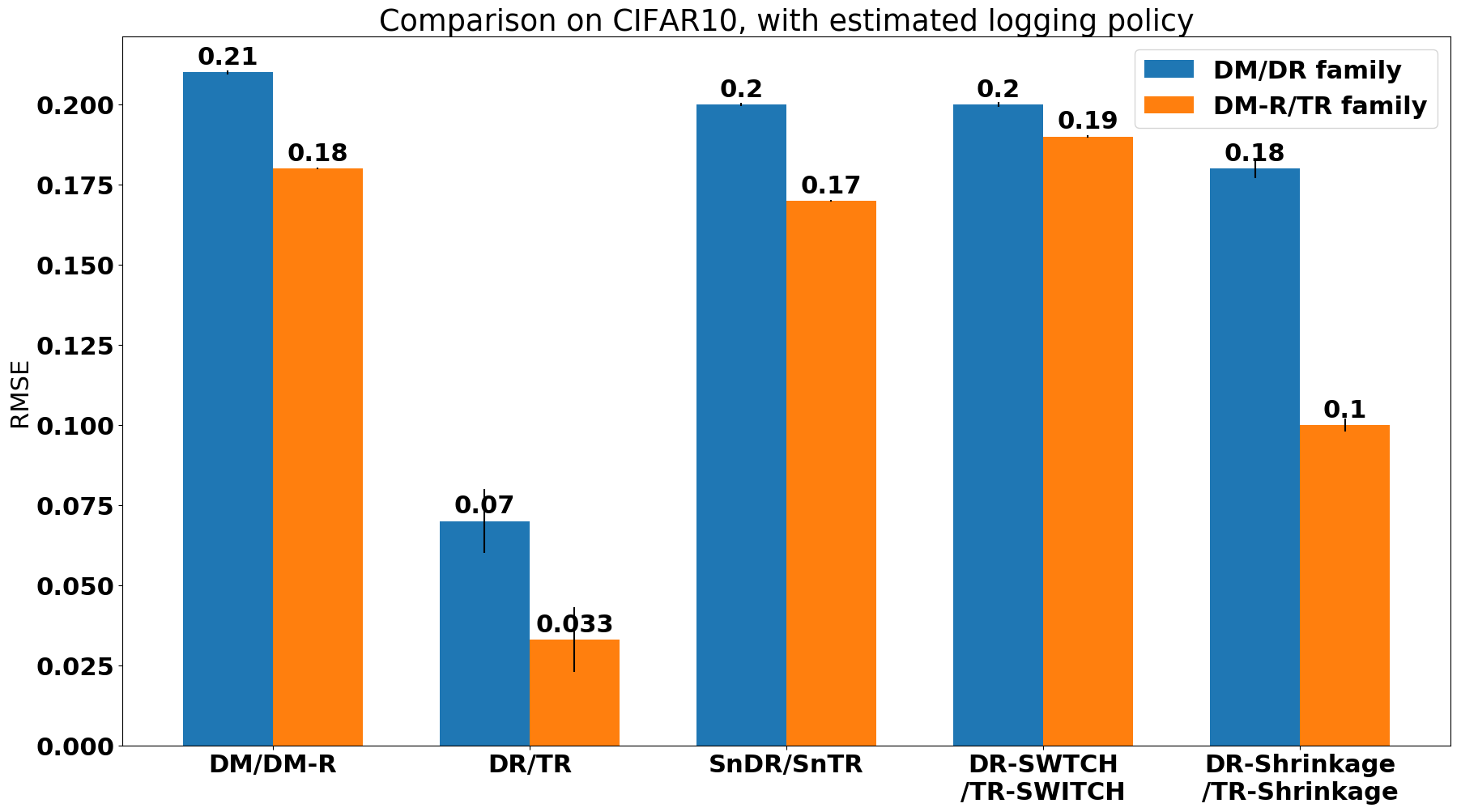}\\
    (a)&(b)
  \vspace{-0.1in}
\end{tabular}
    \caption{(a) Performance Comparison in RMSE on Vehicle, when logging policy is known but with high variance. DR/TR fails due to variance in logging, but DM-R is able to outperform DM, and further improve SnTR and TR-SWITCH over DR counterparts. (b) Performance Comparison in RMSE on CIFAR10 when logging policy is estimated from data.  Augmenting existing methods improves performance across the board. 
    }
\label{fig:exp}
\end{figure*}

\textbf{Methods Compared.}
We provide performance comparison with state-of-the-art methods. We classify the off-policy evaluation methods into two categories: 
\vspace{-0.1in}
\begin{itemize}
\item{\bf DM/IPS/DR family} includes DM \eqref{VDM}, IPS \eqref{VIPS}, SnIPS \eqref{VSnIPS}, DR \eqref{VDR}, DR-SWITCH \eqref{VSWITCH}, DR-Shrinkage \eqref{VDRS}, and SnDR that uses SnIPS in DR. 

\vspace{-0.05in}
\item {\bf DM-R/TR family} includes DM-R \eqref{VDMR}, TR \eqref{VTR}, SnTR \eqref{VSnTR}, and TR-SWITCH \eqref{VTRSWITCH}, and TR-Shrinkage \eqref{VTRS}.

\end{itemize}
We evaluate all the above methods in our experiments. When a reward model is needed, we adopt deep neural networks as representation $f(x, a)$. In SWITCH and Shrinkage, we set a hard threshold as $\tau = 0.5$ or $\hat{w}(x, a) = 0.5$ respectively when it is greater than 0.5. The reason is for a fair comparison with other method that does not require careful hyperparameter search. We report three sets of results where logging policy is obtained differently. Table \ref{tab:main_experiments} top is with known and uniform logging policy, which means $p(a|x) = \frac{1}{\#Classes}$. Table \ref{tab:main_experiments} middle is with known logging policy that is estimate from a biased subsampled data from training. Table \ref{tab:main_experiments} bottom is with estimated logging policy using a classification model. We show the best performing method in each family. To demonstrate how robust regression affects direct method and doubly robust respectively, we also given more detailed comparison for Viehcle in the higher variance logging policy case
and CIFAR10 when using estimated logging policy in Figure \ref{fig:exp}. We only show pairwise comparison for counterparts in DM/DR family and DM-R/TR family.

\subsection{Performance Analysis}
In all the cases, best performing methods in DM-R/TR family outperform the ones in DM/IPS/TR family. Especially, in the challenging case when the logging policy needs to be estimated from data, we achieve a even larger gap from the best performing baseline, as shown in the bottom table in Table \ref{tab:main_experiments}.
We can also observe the following from the experimental results.

{\bf With known and uniform logging policy:} IPS is accurate and small variance in this case, so both DR and TR achieve good results and TR can outperform DR with smaller variance. This is also true for variants methods DR-Shrinkage and TR-Shrinkage. 

{\bf With known but high variance logging policy:} TR outperform baselines most of the time. The only exception is shown in Figure \ref{fig:exp} (a), when logging policy is high variance and DM/DM-R achieves best error. DR/TR suffer from the variance. In this case, DM-R outperforms DM and the benefit directly transfers to variants methods. 

{\bf With estimated logging policy:} Even though the RMSE is generally larger in all methods, comparing against known policy cases, best performing methods in DM-R/TR family still can improve over DM/DR family. Moreover, Figure \ref{fig:exp}(b) shows TR reduce the error by half than DR in cifar10, thanks to a better direct method. 

{\bf Does DM-R always outperform DM and TR always outperform DR?} The answer for former is yes almost all the time. This is due to fact that robust regression considers the shifts explicitly. But the variance of IPS could make both DR and TR suffer, in which case TR-variants or DM-R wins. Therefore, when IPS has low variance, DM-R always help TR and its variants to beat DR counterparts. When IPS has higher variance, DM-R can still outperform DM and transfer its benefit to TR and its variants. 




\section{Conclusion and Future Work}
We propose to use deep robust regression for off-policy evaluation problem under the contextual bandit setting. We demonstrate how it serves a better direct method (DM-R) and also improves all the doubly robust variants when using it in the DM component of DR, which we denote as the Triply Robust (TR) method. We prove novel bias and variance analysis for TR and a minimax bound for DM-R. Experiments demonstrate that DM-R/TR family methods achieve better empirical performance than their counterparts. 

We plan to advance our studies from the following perspectives:
There are several DR methods that we can further improve using robust regression \cite{agarwal2017effective,swaminathan2017off}. We also plan to investigate a more advanced logging policy estimation method and study how it interplay in TR. Finally, how TR can further benefit off-policy reinforcement learning is also a natural next step. Given recent advances in batch reinforcement learning \cite{highConfidence,sdr,magic,farajtabar2018more,fqe,kallus2019intrinsically}, it would be interesting to see how TR methods can interact and compare with them. 






\clearpage
\section*{Acknowledgement}
 Prof. Anandkumar is supported by Bren endowed Chair, faculty awards from Microsoft, Google, and Adobe, DARPA PAI and LwLL grants. Anqi Liu is a PIMCO postdoctoral fellow at Caltech.
\begingroup
\bibliography{bib}

\endgroup
\newpage
\appendix

\section{Derivation of Robust Regression Model}
\label{app_a}
According to~\cite{chen2016robust}, the solution of the minimax formulation has the parametric form: 
\begin{align}
\hat{P}(r|x, a) \propto P_0(r|x, a) e^{\frac{P(x)p(a|x)}{P(x)\pi(a|x)}\theta^{T}\Phi( x, a, r)} \notag\\ 
= P_0(r|x, a) e^{\frac{p(a|x)}{\pi(a|x)}\theta^{T}\Phi( x, a, r)}
\end{align}
The parameters are obtained maximum condition log likelihood estimation with respect to the target distribution: 
$$
    \theta = \arg\max_{\theta}E_{a\sim \pi(a|x), r \sim P(r|x, a)}\left[
    \log\hat{P}_{\theta}(R| X, A)
    \right].
$$
If we set the potential function $\theta \cdot \Phi( x, a, r)$ in a special way such that it has a quadratic form and assuming the base distribution $P_0\sim N(\mu_0, \sigma_0)$, we obtain a Gaussian distribution. We provide necessary details here for the derivation of the mean and variance from this special form and refer more details to \cite{chen2016robust}.
If potential function has this form:
\begin{align}
   \theta \cdot \text{vector}(\Phi( x, a, r)) 
   = \rho \cdot \begin{bmatrix} r \\ f(x, a) \end{bmatrix}^T \begin{bmatrix} r \\ f(x, a) \end{bmatrix},\label{eq:potential}
\end{align}

The optimization of parameters $\rho$ involve maximizing the target loglikelihood:
$\rho = \argmax \mathbb{E}_{a \sim \pi(a|x), r\sim P(r|x, a)}\left[\log\hat{P}_{\rho}(R| X, A)\right]$. The gradients of $\rho_r$ and $\rho_{xr}$ are as follows:
    \begin{align}
        &\nabla_{\rho_r}\mathbb{E}_{a\sim \pi(a|x), r \sim P(r|x, a)}\left[
    \log\hat{P}_{\rho}(R| X, A)
    \right] \notag\\ 
    &=  \frac{1}{n}\sum^n_i r_i^2 - \frac{1}{n}\sum^n_i\mu^2(x_i,a_i, \rho) - \sigma^2(x_i, a_i, \rho), \label{eq:gradients1}\\
    &\nabla_{\rho_{xr}} \mathbb{E}_{a \sim p(a|x), r \sim P(r|x, a)}\left[
    \log\hat{P}_{\rho}(R| X, A)
    \right] \notag\\
    &= \frac{1}{n}\sum^n_i (r_i -  \mu(x_i, a_i, \rho)) f(x, a).\label{eq:gradients2}
    \end{align}
    
\section{Proof of Theorem \ref{thm:bias} and Theorem \ref{thm:variance}}
\label{app_b}
\begin{proof}
We first restate a generalization bound and prove a lemma for robust regression.
\begin{theorem} (of \cite{liu2019robust})
The generalization error of robust regression is upper bounded by the following with probability at least $1-\delta$:
{\small
\begin{align}
  &E_{a\sim \pi(a|x), r\sim P(r|x, a)}[(r - \hat{\mu}_a)]\notag
    \\
    &\le  \sqrt{W}\left[(2WB + \frac{1}{\sigma_0^2})^{-1} + \eta_1 + 4M\hat{\mathfrak{R}}(\mathcal{F})+ 3M^2\sqrt{\frac{\log\frac{2}{\delta}}{2n}}\right]^{\frac{1}{2}} \notag\\
    &:= \epsilon,\label{eq:epsilon}
\end{align}
}
where $W$ is the upper bound of $\frac{\pi(a|x)}{p(a|x)}$, $B$ is the upper bound of model parameter $\rho_r$, $\mathcal{F}$ is the function class of $f$ with $\sup_{x\in \mathcal{X},f,f'\in \mathcal{F}} |f(x) -f'(x)|\le M$ whose Rademacher complexity is $\hat{\mathfrak{R}}$, $\sigma_0^2$ is the base distribution variance, and $n$ is the number of data samples. 
\end{theorem}
Moreover, a property from robust regression is that when the model is fully optimized, we have the following lemma:
\begin{lemma}
The $L_1$ distance between the first and second order of the mean estimators from robust regression satisfies the following with probability at least $1 - \delta$, if both $r$ and $\hat{\mu}_a$ is bounded by 1:
\begin{align}
    &E_{a\sim p(a|x), r\sim P(r|x, a)}[|r - \hat{\mu}_a|] 
    \le \eta_1/l  + O(\sqrt{\log (1/\delta)/n}); 
    \notag\\
    &E_{a \sim p(a|x), r\sim P(r|x, a)}[|r^2 - \hat{\mu}_a^2|] \notag\\
    & \quad\quad\quad\quad\le  \eta_2 + \frac{1}{n}\sum^n_i\hat{\sigma}_a^2 + O(\sqrt{\log (1/\delta)/n});\label{eq:lemma_p}
\end{align}
where $\hat{\sigma}_a^2$ is the variance prediction from robust regression model, $l$ is the lower bound of all the features in context and action pair representation $f(x, a)$, and $n$ is the number of data samples.
\end{lemma}
Therefore, we can use these tools to further analyze bias and variance of the triply robust method.

Due to the theoretical property of robust regression, we have \eqref{eq:lemma_p} and \eqref{eq:epsilon} hold.
Therefore, we have
\begin{align}
     &| E_{a\sim \pi(a|x), r\sim P(r|x,a)}[\hat{V}_{TR}] - V^{\pi}] | \notag \\
     &\le |\mathbb{E}_{a\sim p(a|x)}[W (r - \hat{\mu}_a]| + |\mathbb{E}_{a\sim \pi(a|x)}[(r - \hat{\mu}_{\pi})] | \notag\\
     &\le W\eta_1/l  + \epsilon + O(\sqrt{W\log (1/\delta)/n}).
\end{align}

According to Theorem 2 in \cite{dudik2011doubly}, if logging policy is accurate, the magnitude of variance of DR depends on $E_x[(r_a - \hat{r}_a)^2]$, therefore, in the triply robust case, we have:
\begin{align}
    &E_{a\sim p(a|x), r\sim P(r|x,a)}[(r_a - \hat{\mu}_a)^2] \notag\\
    &\le \eta_2 + \sigma^2(x, a, \rho) + O(\sqrt{\log (1/\delta)/n})
\end{align}

We can plug in this bound to the original DR variance analysis and obtain a new bound. 
Similarly, according to the decomposition of the variance of DR, we have:
\begin{align}
    &VAR_{a\sim \pi(a|x), r\sim P(r|x,a)}[\hat{V}_{TR}] \\
    & \le E_{a\sim p(a|x)} [(W(r_a - \hat{\mu}_a) + r - \hat{\mu}_{\pi})^2 ]\\
    & \le 2\mathbb{E}_{a\sim p(a|x)}[W(r - \hat{\mu}_a)^2]  + 2\mathbb{E}_{a\sim \pi(a|x)}[(r - \hat{\mu}_{\pi})^2]\\
    & \le2 W^2\eta_2 + 2\frac{1}{n}\sum^n_iW^2\sigma^2(x_i, a_i, \rho) \\
    & \quad\quad\quad\quad\quad\quad\quad\quad\quad+ O(W^2\sqrt{\log (1/\delta)/n}) + 2\epsilon^2
\end{align}
According to ~\cite{liu2019robust}, we have $\frac{1}{n}\sum^n_i\sigma^2(x_i, a_i, \rho)\le (2WB + \frac{1}{\sigma_0^2})^{-1}$, where $W$ is the upper bound of $\frac{\pi(a|x)}{p(a|x)}$, $B$ is the upper bound of model parameter $\rho_r$, therefore 
\begin{align}
    &VAR_{a\sim \pi(a|x), r\sim P(r|x,a)}[\hat{V}_{TR}]\notag\\
    &\le 2W^2\eta_2 + 2W^2(2WB + \frac{1}{\sigma_0^2})^{-1}\notag\\
    &\quad\quad\quad\quad\quad\quad\quad\quad\quad + O(W^2\sqrt{\log (1/\delta)/n}) + 2\epsilon^2
\end{align}

\end{proof}
\section{Proof of Theorem \ref{thm:minimax}}
\label{app_c}
\begin{proof}
Robust regression assumes the worst-case data generating distribution that satisfies constraints from feature means of training data. This translates to the mean and variance constraints for the resulting conditional Gaussian distribution. We have the following two lemmas.
\begin{lemma}
\label{lemma:constraint_check}
The max player $\check{P}(r|x,a)$ in the minimax framework of robust regression when feature function take the form of \eqref{eq:potential} satisfies:
{\small 
\begin{dmath}
    \left|E_{a\sim p(a|x),r\sim\check{P}(r|x,a)}[r(x, a)^2] - \frac{1}{n}\sum_i r_i^2 \notag\\
    \quad\quad\quad + \text{VAR}_{a\sim p(a|x), r\sim \check{P}(r|x,a)}[r(x, a)] \right| \le \eta_1 ;\notag\\
    \left|E_{a\sim p(a|x),r\sim\check{P}(r|x,a)}[r(x, a)] - \frac{1}{n}\sum_i r_i \right|f(x, a) 
     \le \eta_2; \label{eq:constraints}
\end{dmath}
}
where $\eta_1$ and $\eta_2$ are the slack we can set for the constraints.
\end{lemma}
\begin{lemma}
\label{lemma:constraint_hat}
The estimator solved from the minimax framework when $\Gamma$ is \eqref{eq:constraints} also satisfies \eqref{eq:constraints}.
\end{lemma}

Satisfying such constraints, we are interested in what is the lowest MSE that any estimator can achieve. 
Denote $\mathcal{S}(\eta_1, \eta_2, f)$ as the set of distributions that satisfy \eqref{eq:constraints}, we are interested in the minimax risk $\mathcal{R}(P, p, \pi, \eta_1, \eta_2, f)$ of off-policy evaluation over the class  $\mathcal{S}(\eta_1, \eta_2, f)$, which is defined as 
\begin{align}
    \inf_{\hat{r}} \sup_{D(r|x, a) \in \mathcal{S}(\eta_1, \eta_2, f)} E_{a\sim \pi(a|x), r\sim D(r|x,a)}[(\hat{r} - r(x, a))^2] \notag
\end{align}
We analyze the minimax risk in terms of the mean squared error, even though we optimize the relative loss \eqref{eq:relativeloss} in practice. Because it is more natural and convenient to obtain uncertainties from the relative loss, which provides significant benefit in practice.
The key of this proof follows the idea of classic minimax theory ~\cite{lafferty9minimax}. We first reduce the problem to hypothesis testing and then pick parameters for the testing to obtain the final bound.

Recall that our setup assumes that the logging policy $p(a|x)$ and the evaluation policy $\pi(a|x)$, the context distribution $P(x)$, the constraints slacks $\eta_1$ and $\eta_2$ fixed. According to Lemma \ref{lemma:constraint_hat}, the resulting conditional Gaussian distribution satisfies constraints in \eqref{eq:constraints}. Then instead of setting a general upper bound for the mean and variance of the ``sup" player in $\mathcal{R}$, we use the constraints in \eqref{eq:constraints} to define a more specific set of distributions where the adversarial player $\hat{P}(r|x, a)$ come from.

We now construct a set of distribution $\mathcal{S}$ that satisfy the constraints. For a given $\eta_1$ and $\eta_2$, the reward distribution $D(r|x, a)$ is a Gaussian distribution with mean $\mu(x, a)$ and variance $\sigma^2(x, a)$ such that they satisfy \eqref{eq:constraints}.

For any two distributions $D_1 (\mathcal{N}(\mu_1(x, a), \sigma_1^2(x, a)))$ and $D_2 (\mathcal{N}(\mu_1(x, a), \sigma_1^2(x, a)))$ in $\mathcal{S}$, we have the lower bound:
\begin{align}
    \mathcal{R} \ge \inf_{\hat{v}} \max_{D\in D_1, D_2} E_{D}[(\hat{v} - v_D)^2].
\end{align}
Here we use subscription $D$ to represent data distribution that contexts and actions are drawn based on $P(x)$ and $p(a|x)$ and the reward is drawn from $D(r|x, a)$.
For a $\lambda$ to be chosen later, we have the following:
\begin{align}
    \mathcal{R} \ge \max_{D\in D_1, D_2} E_D[(\hat{v} - v_D)^2] \ge \max_{D\in D_1, D_2}\lambda P_D((\hat{v} - v_D)^2) \notag\\
    \ge \frac{2}{\lambda}\left[ P_{D_1}((\hat{v} - v_{D_1})^2\ge \lambda) + P_{D_2}((\hat{v} - v_{D_2})^2\ge\lambda)\right].\label{eq:testing}
\end{align}
For turning the problem to a testing problem, the idea is to identify a pair of distribution $D_1$ and $D_2$ such that they are far enough from each other so that any estimator which gets a small estimation loss can essentially identify whether the data generating distribution is $P_{D_1}$ or $P_{D_2}$. In order to do this, we take any estimator $\hat{v}$ and identify a corresponding test statistic which maps $\hat{v}$ into one of $D_1$ and $D_2$. The way to do this is identified in \eqref{eq:testing}. 

For any estimator $\hat{v}$, we can associate a statistic $s(\hat{v}) = \argmin_{D} \{(\hat{v} - v_{D_1})^2, (\hat{v} - v_{D_2})^2 \} $. Therefore, we are interested in its error rate $P_D(s(\hat{v}) \neq D)$. We can prove that if $(v_{D_1} - v_{D_2})^2 \ge 4\lambda$, it yields \eqref{eq:testing}.

Now we place a lower bound on the error of this test statistic. Using the result of Le Cam \cite{lafferty9minimax}, which places an upper bound on the attainable error in any testing problem. This translate to the following in our problem:
\begin{align}
    \max_{D\in D_1, D_2}P_{D}(s(\hat{v}) \neq D) \ge \frac{1}{8}e^{-n D_{KL}(P_{D_1}||P_{D_2})}.
\end{align}
Since we would like the probability of error in the test to be a constant, it suffice to choose $D_1$ and $D_2$ such that 
\begin{align}
    D_{KL}(P_{D_1}||P_{D_2}) \le \frac{1}{n}.\label{eq:kl_constraints}
\end{align}

We next make concrete choices for $D_1$ and $D_2$. The constraints we need to satisfy are \eqref{eq:kl_constraints}
and \eqref{eq:constraints}, which ensure that $D_1$ and $D_2$ are not too close that an estimator does not have to identify the true parameter, or too far that the testing problem becomes too trivial. In order to find a reasonable choice of $D_1$ and $D_2$, we assume $\mu_1 = E_{a\sim p(a|x),r\sim\check{P}(r|x,a)}[r(x, a)]$ and $\mu_2 = E_{a\sim p(a|x),r\sim\check{P}(r|x,a)}[r(x, a)] + \alpha$. Then we have $\alpha \le \eta_2/l$. According to \eqref{eq:constraints}, we construct the following variances for the Gaussian distribution:
\begin{align}
\sigma^2_1 = \sigma^2_2 \le \eta_1 - \alpha^2 - 2\alpha E_{a\sim p(a|x),r\sim\check{P}(r|x,a)}[r(x, a)].
\end{align}
This construction makes sure both $D_1$ and $D_2$ satisfy \eqref{eq:constraints}. From now on we just use $\sigma^2$ to represent the variance.

Since both distribution of rewards is a Gaussian and they have the same variance now. The KL-divergence is given by the squared distance between the means, scaled by the variance, which is:
\begin{align}
    D_{KL}(P_{D_1}||P_{D_2}) = E\left[\frac{(\mu_1(x,a) - \mu_2(x, a))^2}{\sigma^2} \right]
\end{align}
Thus we have:
\begin{align}
    E\left[\frac{\alpha^2}{2\sigma^2}\right] \le \frac{1}{n}.
\end{align}
The minimax lower bound is then obtained by the largest $\alpha$ in such that the other constraints are satisfied. This gives the following optimization problem:
\begin{align}
    &\max \lambda\\
    s.t. \quad\quad&E[w \alpha] \ge 2\sqrt{\lambda},\\
    &E\left[\frac{\alpha^2}{2\sigma^2}\right]\le \frac{1}{n},\\
    &\alpha \le \eta_2/l\\
    &\sigma^2 \le  \eta_1 - \alpha^2 - 2\alpha E_{a\sim p(a|x)}[r(x, a)].
\end{align}
Solving for $\lambda$, we have 
\begin{align}
    &\alpha = \min\Big\{ \frac{\eta_2}{l}, \\
    &\quad\quad\frac{-4E_{ p}[r] + \sqrt{16 E_{p}[r] ^2+ 8(n+2)\eta_1}}{2(n+2)} \Big\},
\end{align}
where we use $E_p$ to represent $E_{a\sim p(a|x),r\sim P(r|x,a)}$.
If we have $E_{p}[w\alpha] =: 2\sqrt{\lambda}$, putting together the bounds together, we obtain:
\begin{align}
    &\mathcal{R} \ge \frac{\lambda}{2} \cdot(\max_{D\in D_1, D_2}P_{D}(s(\hat{v})) \neq D)\\
    & \ge \frac{\lambda}{2}\cdot \frac{1}{8}e^{-n D_{KL}(P_{D_1}|| P_{D_2})}\\
    & \ge \frac{\lambda}{16e}\\
    & \ge \frac{E[w\alpha]^2}{64e}
\end{align}
Therefore, the lower bound of $\mathcal{R}$ is,
\begin{align}
        &\alpha = \min\Big\{ \frac{w^2\eta^2_2}{64el^2}, \\
    &\quad\quad\frac{(-4E_{p}[wr] + \sqrt{16 E_{p}[wr] ^2+ 8w^2(n+2)\eta_1})^2}{128e(n+2)^2} \Big\},
\end{align}
\end{proof}

\section{More method implemented}
\label{app_d}
We also implemented a version of direct method that use an ablation version of DM-R. The robust regression framework is not limited to the covariate shift case. An observation is that if $\frac{p(a|x)}{\pi(a|x)} = 1$ for all the actions, which means there is no covariate shift or we ignore the shift, we obtain a version of robust regression that can be applied to i.i.d. data.  In this case, we robustly minimize the relative loss under that data distribution generated by logging policy: $\mathcal{L}':=E_{a\sim p(a|x), r\sim P(r|x, a)}\left[
    -\log\frac{\hat{P}(R| X, A)}{P_0(R|  X, A)}
    \right]$, which has the solution as a Gaussian distribution with mean and variance as follows:
    \begin{align}
    \mu'(x,a, \rho') =&
 \left(2 \rho_r' + \sigma_0^{-2}\right)^{-1}\left(-2 \rho'_{xr} 
    f(x, a)
    + \mu_0\sigma_0^{-2}\right),\notag\\
    \sigma^{2'}(x, a, \rho') = &\left(2 \rho_r' + \sigma_0^{-2}\right)^{-1}.\label{DMI}
    \end{align}

We then plug in the mean estimates from \eqref{DMI} in direct method and obtain:
\begin{eqnarray}
  \hat{V}_{\text{DM-I}} = \frac{1}{|S|}\sum_{x \in S}E_{a\sim\pi(x)}\left[\mu'(x,a, \rho')\right].\label{VDMI}
\end{eqnarray}

\section{Experimental details}
\label{app_e}
The network structure we used in experiments is a 4 layer fully-connected spectral normalized one, with 64 hidden nodes for UIC datasets. Resnet18 is used for CIFAR10. The training epochs for generating the ground truth is set to be 5. The training epochs for training reward models are set to be 20. The learning rate for stochastic gradient descent is set to be 0.0001. In the prediction, we round the regression result to be within $[0, 1]$. The base distribution for robust regression is a Gaussian distribution with mean 0.5 and variance 1, since we know the reward is between $[0, 1]$. 

For the known logging policy case, except for the uniform logging policy case, we need to train a probabilistic model from a subsampled dataset. We use the same NN architecture with the reward model and train on fully observed data to obtain a ``sample model". We can control the variance by setting different regularization values and training epochs. Similarly, when we need to estimate the logging policy, we use the data generated by the ``sample model" as training data to train a ``policy model". 

\section{More Experimental Results}
We put the full experimental results in Table \ref{tab:details1} and Table \ref{tab:details2}.
\begin{table*}[htb]
    \centering
     \caption{Comparing DM/IPS/DR family with their performance in RMSE mean and standard deviation (in parentheses) over 20 repeated experiments. }
    \begin{tabular}{|K{1.2cm}|K{1.5cm}|K{1.5cm}|K{1.5cm}|K{1.5cm}|K{1.5cm}|K{1.5cm}|K{1.5cm}|K{1.5cm}|}
         \hline
         Data&  policy &DM & IPS & SnIPS &DR & SnDR & DR-SWITCH & DR-Shrinkage\\
         \hline
         \multirow{3}{4em}{optdigits} &uniform &0.24 (10e-6)  & 0.064 (0.035) & 0.62 (10e-6) & 0.046 (0.03) & 0.24 (10e-6) & 0.24 (10e-6) & 0.18 (0.006) \\
         \cline{2-9}
         &biased &0.40 (10e-6) & 0.24 (0.23) & 0.55 (10e-6) & 0.21 (0.20) &0.40 (10e-6)& 0.41 (0.0015) & 0.42 (0.0048) \\
         \cline{2-9}
         &estimated &0.59 (10e-6)& 0.58 (0.027) & 0.67 (10e-6) & 0.53 (0.025) & 0.60 (0.00054) & 0.61 (0.027) & 0.61 (10-6)\\
         \hline
         \multirow{3}{4em}{vehicle} &uniform &0.06 (10e-6)  &0.07 (0.063)  & 0.70 (10e-6) &  0.05  (0.047) & 0.064 (0.00027) & 0.065 (10e-6)& 0.027 (0.024) \\
         \cline{2-9}
         &  biased&0.07 (10e-6)& 0.16 (0.0056)  & 0.80 (10e-6) &  0.14 (0.034)& 0.07 (0.0001) & 0.07 (0.0003) & 0.14 (0.022)  \\
         \cline{2-9}
         & estimated&0.75 (10e-6) & 0.21 (0.08) &0.75 (0.0004) & 0.25 (0.12) & 0.75 (0.0005) & 0.75 (10-6) & 0.60 (0.024)\\
         \hline
        
         \multirow{3}{4em}{letter} &uniform&0.13 (10e-6)& 0.030 (0.0027)& 0.81 (10e-6) & 0.21 (0.0021)  & 0.13     (10e-6) & 0.13 (10e-6) & 0.12 (0.0016)  \\
         \cline{2-9}  
         &biased&0.20 (10e-6) & 0.15 (0.015)&  0.89 (10e-6)& 0.06 (0.061)  & 0.20 (10e-6) & 0.22 (0.00066)  & 0.22 (0.0023)\\
         \cline{2-9}
         & estimated&0.24 (10e-6) & 0.05  (0.026)& 0.48 (10e-6)&  0.033 (0.015)& 0.24 (10e06) & 0.18 (0.0012) & 0.06 (0.0043) \\
         \hline
          \multirow{3}{4em}{cifar10} &uniform&0.27 (0.0004)& 0.01 (0.012)& 0.62 (10e-6) & 0.012 (0.0008)  & 0.27 (0.0004) & 0.27 (0.0004) & 0.21 (0.0016)   \\
         \cline{2-9}  
         &biased&0.18 (0.00032) & 0.018 (0.032)&  0.60 (10e-6)& 0.015 (0.0050)  & 0.18 (0.00032) & 0.16 (0.00028) & 0.084 (0.0021) \\
         \cline{2-9}
         & estimated&0.18 (0.00063) & 0.10 (0.0080)& 0.61 (10e-6)&  0.07 (0.012)& 0.18 (0.000063) & 0.18 (0.00086)& 0.084 (0.0028) \\
        
         \hline
    \end{tabular}
    \label{tab:details1}
\end{table*}

\begin{table*}[htb]
    \centering
     \caption{Comparing DM-R/TR family with their performance in RMSE mean and standard deviation (in parentheses) over 20 repeated experiments. }
    \begin{tabular}{|K{1.2cm}|K{1.5cm}|K{1.5cm}|K{1.8cm}|K{1.8cm}|K{1.8cm}|K{1.8cm}|K{1.9cm}|}
         \hline
         Data&  policy &DM-R & DM-I &  TR& SnTR & TR-SWITCH & TR-Shrinkage\\
         \hline
         \multirow{3}{4em}{optdigits} & uniform&0.23 (10e-6)  & 0.027 (0.017) & 0.045 (0.028) & 0.23 (10e-6) & 0.23 (10e-6) & 0.17 (0.0056) \\  
         \cline{2-8}
         & biased&0.28 (10e-6) & 0.41 (10e-6) & 0.13 (0.13) & 0.28 (10e-6) &0.29 (0.0013)&0.27 (0.0040) \\ 
         \cline{2-8}
         & estimated &0.55 (10e-6)& 0.54 (10e-6) & 0.47 (0.02) & 0.55 (10e-6) & 0.56 (0.00055) & 0.54 (0.0027)\\ 
         \hline
         \multirow{3}{4em}{vehicle} & uniform&0.066 (10e-6)  &0.38 (10e-6)  & 0.060 (0.045) &  0.065 (0.00026) & 0.066 (10e-6) & 0.026 (0.022) \\ 
         \cline{2-8}
         &  biased&0.0076 (10e-6)& 0.50 (10e-6)  & 0.15 (0.033) &  0.0080 (0.00010)& 0.011 (0.0027) & 0.15 (0.023)\\ 
         \cline{2-8}
         & estimated&0.24 (10e-6) & 0.24 (10e-6) &0.24 (0.23) & 0.24 (0.00024) & 0.24 (10e-6) & 0.18 (0.013) \\ 
         \hline
         \multirow{3}{4em}{letter} &uniform& 0.054 (10e-6)& 0.099 (10e-6)& 0.019 (0.019) & 0.054 (10e-6) &0.054 (10e-6) & 0.050 (0.0014)  \\ 
         \cline{2-8}  
         &biased&0.22 (10e-6) & 0.20 (10e-6)&  0.039 (0.069)& 0.22 (10e-6)  & 0.23 (0.00040) & 0.24 (0.0024)\\ 
         \cline{2-8}
         & estimated&0.22 (10e-6) & 0.17 (10e-6)& 0.023 (0.015)&  0.22 (10e-6)& 0.18 (0.0010) & 0.069 (0.0043) \\ 
        
         \hline
         \multirow{3}{4em}{cifar10} &uniform&0.24 (0.00014)& 0.29 (10e-6)& 0.011 (0.0092) & 0.24 (0.00014)  & 0.24 (0.00014) & 0.19 (0.0017) \\ 
         \cline{2-8}  
         &biased&0.21 (10e-6) & 0.20 (0.00018)&  0.012 (0.0060)& 0.21(10e-6)  & 0.20 (0.00026) & 0.10 (0.0023)\\ 
         \cline{2-8}
         &estimated &0.22 (0.00016) & 0.18 (0.00026)& 0.033 (0.012)&  0.22 (0.00016)& 0.21 (0.00043)&0.10 (0.0027) \\
        
         \hline
    \end{tabular}
    \label{tab:details2}
\end{table*}

\end{document}